\documentclass{article}

\usepackage{preprint}
\usepackage[utf8]{inputenc}
\usepackage{url}
\usepackage{amsmath}
\usepackage{amsfonts}
\usepackage{nicefrac}
\usepackage{microtype}
\usepackage{xcolor}
\usepackage{graphicx}
\usepackage{booktabs}
\usepackage{float}   % [H]: pin the appendix provenance table in place
\title{Real-Time Detection and Repair of LLM Agent Failures}

\author{
  Sunny Dubey \\
  ORCID: \texttt{0009-0002-8296-8631} \\
}

\begin{document}

\maketitle

\begin{abstract}
LLM agents fail mid-episode---they loop, cascade tool errors, drift off their goal, fabricate results, or silently absorb corrupted content---and the standard remedy, judging every step with a second LLM, costs more than the agent itself. We ask how much failure detection is achievable from \textit{observable step telemetry alone} (semantic embeddings of step output, token-level uncertainty, action metadata), using monitors that cost microseconds per step and train only on healthy runs. Starting from a one-class echo-state-network (ESN) ensemble with CUSUM alarms, we build and validate, on 2,823 committed agent episodes across three frameworks, three local agent models spanning two families (qwen2.5 7b/3b, llama3.1 8b), and a commercial API (gemini-2.5-flash), a sequence of increasingly capable monitors: (1) the ESN, which wins decisively when failures have temporal room to develop; (2) a calibrated ESN+Mahalanobis hybrid whose learned fusion weights track the deployment regime; and (3) a content-grounding telemetry channel that lifts the monitors' shared blind spot (content corruption) from 0.28 to 0.59 pooled content detection (reaching 1.00 on the research collectors where the corruption is content-visible; honestly inert on frameworks whose corruption leaves result text unchanged), a $+0.31$ gain that holds at every seed, while behavioral detection improves rather than degrades ($+0.045$). Validation on organic (non-injected) failures reveals a taxonomy with an honest boundary: a completion check catches failures of omission (7/7 silent aborts), the monitors transfer only \emph{weakly} to organic commission (1 of 3 fabrications) and rank the organic set at/below chance without recalibration, and plausible-value corruption requires external reference (the escalation layer's job). A pre-registered fabrication study is explicitly underpowered (2 hallucinations in the pre-registered 55 episodes, 9 across all 175 organic episodes and only 2 of those fabricated \emph{inputs}, against a pre-registered minimum of 10) and makes no detection claim --- the models rarely invent --- so the fabrication class is covered by a deterministic numeric-grounding verifier rather than a statistical monitor. Both burdens --- the per-deployment null and the residual false-alarm rate --- motivate a complementary layer that carries neither: \textbf{deterministic verification}, which recomputes a run's stated total from the tool results that run actually received and confirms every required call was made, needing no null, no threshold and no calibration. Head-to-head on the same episodes and labels it catches 60\% of failures (96\% with the coverage check) at \textbf{0 of 63 false positives} against the monitor's 54\% at 17\%; it replicates on 120 later episodes at disjoint seeds scored frozen (54\%, 93\% with coverage, 0 of 64), transfers unchanged across model families to llama3.1:8b (\textbf{110 of 110 at 0 of 10}), and catches \textbf{26 of 26} provoked fabrications --- the class on which a one-class monitor structurally cannot be scored. A third check validates tool results against the shapes their tool can return, flagging \textbf{0 of 1825} healthy episodes while catching 46\% of injected context corruption, 215 of 218 within one step of onset. Detection is then closed into \textbf{repair}: each flagged run is rolled back to its last fact-gathering step and re-run live, recovering \textbf{45\%} of failures against a 16\% resampling control ($p=0.0005$) and lifting net task success from \textbf{52\% to 73\%} for about one extra model call per run. The monitor scores a step in $\sim$200 $\mu$s and the telemetry adapter that feeds it adds a median 674 $\mu$s, so the whole path stays under a millisecond per step --- against one 7B forward pass for an auditor that judges every step.
\end{abstract}

\section{Introduction}

An agent episode is a sequence of steps; each step emits observable telemetry: what the agent said, how confident its tokens were, what tools it called, what they returned, how long everything took. The monitoring question is whether a lightweight, always-on watchdog over this stream can raise a calibrated alarm at derailment onset---steps before the task fails or the budget burns---without model internals, without labels, and without a second LLM in the loop \cite{zheng2023judging}.

This work answers that question constructively, and documents with equal care what such a watchdog cannot do. Contributions:

\begin{enumerate}
    \item \textbf{Real-ecosystem validation at scale, and it is the paper's spine}: 2,823 episodes over 25 datasets, of which 770 use real tools with live agents across bespoke, LangGraph and AutoGen loops, spanning three local agent models in two families and the gemini-2.5-flash API. Per-class coverage across nine real corpora is reported in full, including where it is poor.
    \item \textbf{A controlled testbed with ground-truth onsets} (failure injector, 43-dim telemetry) and an 11-monitor comparison under matched false-alarm budgets, used to expose detector \emph{mechanics} --- ground-truth onsets at arbitrary horizons, and classes the real corpora do not contain. Its telemetry is constructed, so no deployment claim rests on it.
    \item \textbf{A diagnosis of when temporal monitoring pays}: the ESN's detection advantage over a 50$\times$-cheaper Mahalanobis baseline is a monotone function of post-onset horizon.
    \item \textbf{A calibrated hybrid} whose supervised fusion weights learn which regime a deployment is in.
    \item \textbf{A content-grounding telemetry channel} that closes the content blind spot, with behavioral detection improved rather than traded: pooled over the 58 \texttt{malformed\_json} episodes in \texttt{grounding\_diagnosis.csv}, detection rises from 0.07 for the memoryless parent (0.14 for the ESN) to 0.90 for the content gate.
    \item \textbf{Fusion lessons}: per-stream ``dual-budget'' thresholds with a train-max trip restore detections destroyed by shared thresholds.
    \item \textbf{Organic-failure validation with a pre-registered replication}: monitors calibrated on healthy runs and then applied unchanged catch only \textbf{1 of 3} organic fabrications and rank the organic failure set at/below chance (within-organic AUROC 0.31--0.42), while a temperature-matched null still false-alarms on 36\% of healthy episodes --- so a dedicated pre-registered fabrication study is explicitly \textbf{underpowered} (2 genuine hallucinations in 55 episodes vs a pre-registered minimum of 10) and makes no detection claim, and the fabrication class is instead delivered by a deterministic numeric-grounding verifier. Monitors are blind to silent aborts (a trivial completion check catches 7/7), and the healthy null is highly specific to the deployment configuration.
\end{enumerate}

\section{Related Work}

\textbf{Foundations.} Our alarms are classical sequential change detection: CUSUM \cite{page1954cusum} over per-channel surprise, in the framework of abrupt-change detection \cite{basseville1993detection}. The monitors are one-class \cite{chandola2009anomaly} --- fit on healthy episodes only, no failure labels --- and the baselines follow that tradition: Mahalanobis detectors from out-of-distribution detection \cite{lee2018simple,hendrycks2017baseline} and isolation forests \cite{liu2008isolation}. The temporal backbone is an echo-state network \cite{jaeger2001echo}; reservoir computing \cite{lukosevicius2009reservoir} gives an untrained recurrent feature map whose readout fits in closed form, which is what makes the per-deployment recalibration that \S\ref{sec:real} shows to be mandatory affordable at all. Calibration is evaluated with expected calibration error \cite{guo2017calibration}.

\textbf{Agent failure, and detecting it online.} Agent loops interleave reasoning and tool calls \cite{yao2023react} and ship through frameworks such as AutoGen \cite{wu2023autogen}; failure is common and heterogeneous \cite{liu2024agentbench}, and MAST makes that concrete with 14 empirical failure modes over 1600+ annotated traces \cite{cemri2025why} --- our four injected classes are a deliberately narrow, mechanically-verifiable slice of it. Post-hoc attribution asks which agent and step caused a failure and finds it hard (53.5\% on the agent, 14.2\% on the step) \cite{zhang2025which}; ours is the online, pre-outcome question. That position is occupied: AgentForesight audits each prefix with a 7B LLM and releases AFTraj-2K \cite{zhang2026agentforesight}, weakly supervised early alerting learns turn-level risk from trajectory labels alone \cite{baidya2026sparse}, and rule- or model-based runtime guards \cite{wang2025agentspec,wang2025probguard} intervene at the action boundary but need a specification of ``unsafe'' that we do not. Closest of all is PrefixGuard \cite{huang2026prefixguard}, which also scores prefixes online with \emph{no deployment-time LLM call} --- so cheap online monitoring is not by itself our contribution. It induces typed-step adapters offline and then trains a prefix-risk scorer \emph{supervised on terminal outcomes}, which is exactly the resource we do not assume. Nor is the label-free setting itself unoccupied: Trajectory Guard \cite{advani2026trajectoryguard} detects agent-trajectory anomalies in real time from normal behaviour, without anomaly labels, using a Siamese recurrent autoencoder over pre-trained embeddings. \textbf{We therefore claim neither mid-episode detection, nor observable telemetry, nor per-step cheapness, nor label-free training as novel in isolation.} What we claim is the combination and what it costs to run: a one-class monitor that is \emph{causal per step} rather than scoring a completed plan, fitted in 1.7\,s by closed-form ridge regression on healthy runs alone, over a \emph{deterministic hash} embedding that requires no encoder to download or version --- $\sim$200\,$\mu$s per step, no rule set, no outcome supervision, and evaluated against corpora we did not build (\S\ref{sec:external}).

\textbf{Repairing a failed trajectory.} Detection is only worth its cost if something acts on it, and that closing step has its own literature. AgentTether \cite{zhao2026agenttether} is the closest: it localises failure-critical subtrajectories using an \emph{offline normal-behaviour model} much as we fit a healthy null, converts the localisation into behaviour-scoped guidance, and repairs 59\% of failed $\tau$-bench Banking tasks. Two differences set our \S\ref{sec:verify} apart, and neither is an advantage claim. Theirs diagnoses \emph{after} a run has failed and re-executes with guidance; ours alarms mid-episode and rolls back to a checkpoint before the answer is delivered. And their guidance is derived from a localised cause, where our strongest rung deliberately supplies \emph{less} --- the name of the failing check and no values --- which is the finding we did not expect. The numbers are not comparable: different agents, tasks and failure populations, and we make no head-to-head claim.

\textbf{Judges, hallucination, and internals --- the three things we are not.} LLM-as-judge \cite{zheng2023judging,gu2024survey} is the dominant pattern and costs a model call per check, and judges are not neutral instruments \cite{shi2024judging}; we use one as an escalation layer rather than a default, and \S\ref{sec:real} replaces the operating point that analysis had \emph{stipulated} with a measured one. Hallucination detection \cite{ji2023survey,lin2025hallucination} typically needs sampling or model access --- self-consistency \cite{manakul2023selfcheckgpt}, semantic entropy \cite{farquhar2024detecting}, hidden-state probes \cite{kossen2024semantic} --- where a tool-using agent admits a deterministic route: every figure asserted must trace to a tool result received (\S\ref{sec:organic}). Internal-state monitoring is the sharpest challenge to our premise: residual-stream probes detect concealed deception \cite{goldowskydill2025detecting}, and probes predicting eventual failure from the first round report substantially earlier detection than observable behaviour allows \cite{ruan2026doomed}. We do not dispute it --- with the weights in hand, activations beat telemetry. The premise here is the deployment where you do not hold them, and there the choice is not activations versus telemetry but telemetry versus nothing.

\section{Problem and Monitor}
\label{sec:method}

An episode is a step sequence $t=1,\dots,T$. Each step emits an observable
vector built from three causal channels --- a semantic embedding of the step's
output, token-uncertainty aggregates, and action metadata:
\begin{equation}
x_t \;=\; \big[\,e_t \,;\, u_t \,;\, m_t\,\big] \in \mathbb{R}^{d},
\qquad d \in \{43, 51, 60\}.
\label{eq:telemetry}
\end{equation}
An episode is healthy, or contains an onset at unknown step $\tau$ after which
the trajectory distribution shifts and the run ends in failure. From
\emph{healthy episodes only} we want a causal score $s_t = f(x_1,\dots,x_t)$ and
an alarm time $\hat{\tau} = \min\{t : s_t > \theta\}$ maximising detection at a
fixed false-alarm budget, with per-step compute far below one model call.

\textbf{Reservoir.} Per channel $c$, a sparse random recurrent map is fixed at
initialisation and never trained; only a ridge readout $A^{(c)}$ is fitted, on
healthy runs, to predict the next step:
\begin{align}
h^{(c)}_t &= (1-\alpha)\,h^{(c)}_{t-1} + \alpha \tanh\!\big(W^{(c)} h^{(c)}_{t-1} + W^{(c)}_{\mathrm{in}} x^{(c)}_t\big), \label{eq:reservoir}\\
\hat{x}^{(c)}_{t+1} &= A^{(c)}\big[\,h^{(c)}_t \,;\, x^{(c)}_t \,;\, 1\,\big]. \label{eq:readout}
\end{align}
Because $W^{(c)}$ is frozen, \eqref{eq:readout} is a closed-form least-squares
solve --- the reason a fit costs 1.7\,s where a GRU costs 68\,s.

\textbf{Surprise and accumulation.} The per-step surprise is the normalised
one-step prediction error, averaged over an ensemble of $K$ reservoirs, with
$\sigma_{\mathrm{err}}$ the per-dimension residual scale measured on held-out
healthy runs. Slow drift keeps each step locally predictable, so a
short-memory statistic never crosses threshold; a one-sided CUSUM
\cite{page1954cusum} integrates small persistent shifts instead:
\begin{align}
q^{(c)}_t &= \frac{1}{K}\sum_{k=1}^{K}\ \operatorname*{mean}_{d}
  \left(\frac{\hat{x}^{(c)}_{t\,|\,k} - x^{(c)}_t}{\sigma^{(c)}_{\mathrm{err}}}\right)^{2}, \label{eq:surprise}\\
S^{(c)}_t &= \max\!\big(0,\; S^{(c)}_{t-1} + z\big(q^{(c)}_t\big) - \kappa\big),
  \qquad S^{(c)}_0 = 0. \label{eq:cusum}
\end{align}

\textbf{Fusion and threshold.} A shift confined to the 4-dimensional
uncertainty channel is diluted if it is averaged across all $d$ dimensions, so
each channel is accumulated separately and the alarm reads the loudest:
\begin{equation}
s_t \;=\; \max_{c \,\in\, \{e,\,u,\,m\}} S^{(c)}_t,
\qquad
\theta \;=\; Q_{1-\beta}\Big(\big\{\textstyle\max_t s_t \;:\; \text{healthy val episodes}\big\}\Big),
\label{eq:fusion}
\end{equation}
with $\beta$ the false-alarm budget. Every quantity in
\eqref{eq:telemetry}--\eqref{eq:fusion} at step $t$ depends on $x_1,\dots,x_t$
and fit-time constants only, so the monitor is causal by construction; $\theta$
is read from healthy validation episodes and never from test data. \S\ref{sec:sim}
shows the per-channel max in \eqref{eq:fusion}, not the reservoir, carries most
of the margin --- giving a GRU the same wrapper lifts it past our ESN --- which
makes the wrapper the transferable contribution and the reservoir the cheap way
to pay for it.

\textbf{Baselines and variants}, all fitted and scored under this same one-class
causal protocol: memoryless Mahalanobis and $\Delta$-Mahalanobis
\cite{lee2018simple}, cosine and self-drift, rolling entropy, linear AR, and
trained GRU/LSTM/TCN sequence models; hybrids that robust-$z$ calibrate the ESN
and Mahalanobis streams and fuse them by weight, max or logistic regression; a
grounded stream that joins \eqref{eq:fusion} as a memoryless robust-$z$ max over
the content channel $g$ (\S\ref{sec:grounding}); and a five-line completion check
that asks only whether the final step is empty or a bare tool call.

\section{Telemetry}

Each step $t$ emits $x_t$, built in four causal generations:

\begin{table}[htbp]
\centering
\caption{Telemetry vector evolution across versions.}
\label{tab:telemetry}
\begin{tabular}{lp{1cm}p{9.4cm}}
\toprule
Version & Dims & Content \\
\midrule
v1 (base) & 43 & $e_t$: 32-dim hash embedding; $u_t$: 4 token-surprisal aggregates; $m_t$: action one-hot, log latency, outlen, error \\
v2 & 43 & Tool RESULTS appended into step text so corrupted results reach semantic channel \\
v3 (+x) & 51 & 8 derived behavior dims: cosine drift, task-anchor sim, tool success, retry, latency, context ratio, depth, self-cons. \\
v4 (+g) & 60 & 9 content-grounding dims: query-result dissim, result self-cons, JSON-prefix validity, char anomaly, lexical flag \\
\bottomrule
\end{tabular}

\vspace{0.4em}
{\footnotesize Run each as \texttt{py -m <package>.<module>}: the study runners live under \texttt{derail.experiments}, the analyses under \texttt{experimental} and \texttt{verification}.}
\end{table}

The embedding is a deterministic char-3-gram feature hash (no model, no network). v4 telemetry costs a median 674 $\mu$s/step at the adapter (mean 702, p95 1045, over 491 timed steps); the lexical flag alone costs 2 $\mu$s/result.

\section{Real-Ecosystem Validation: the Empirical Spine}
\label{sec:real}

\textbf{Everything this paper claims about deployment rests on this section.} The committed corpus spans 2,823 episodes over 25 datasets --- three frameworks, three local agent models across two families, and the gemini-2.5-flash API --- every episode a live agent run against real tools, with telemetry measured at the adapter rather than constructed.

\begin{figure}[tbp]
\centering
\includegraphics[width=0.95\linewidth]{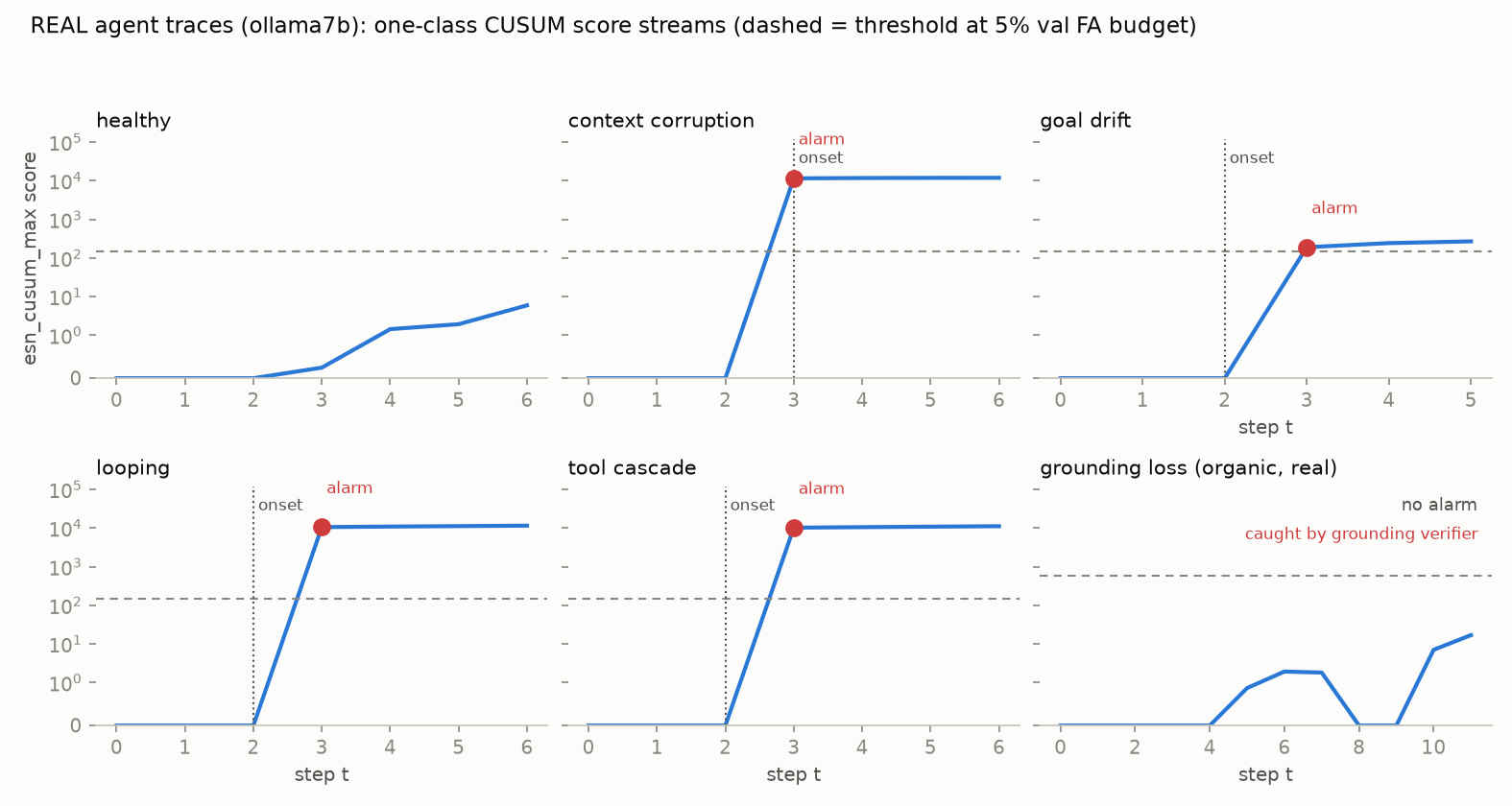}
\caption{\textbf{Real agent traces} (\texttt{ollama7b}: qwen2.5:7b against real tools). One-class CUSUM score streams per failure class; dashed = threshold at the 5\% validation FA budget, dotted = verified injection onset. The healthy run stays two orders of magnitude below the line, and every injected class --- context corruption, \emph{goal drift}, looping and tool cascade --- alarms one step after onset. The sixth panel is \emph{grounding loss}, and it is different in kind: fabrication cannot be injected into a live run, so this episode is a genuine one from the organic (non-injected) corpus, scored against that corpus's own healthy null. The CUSUM stays flat -- a fabricated figure perturbs no behavioural channel -- and the class is caught instead by the deterministic grounding verifier. That is the blind spot the verifier exists to cover, shown rather than asserted. Note the $y$ axis is symlog: real streams span twelve orders of magnitude, because the CUSUM accumulates multiplicatively once a failure takes hold.}
\label{fig:traces}
\end{figure}

\begin{figure}[tbp]
\centering
\includegraphics[width=0.98\linewidth]{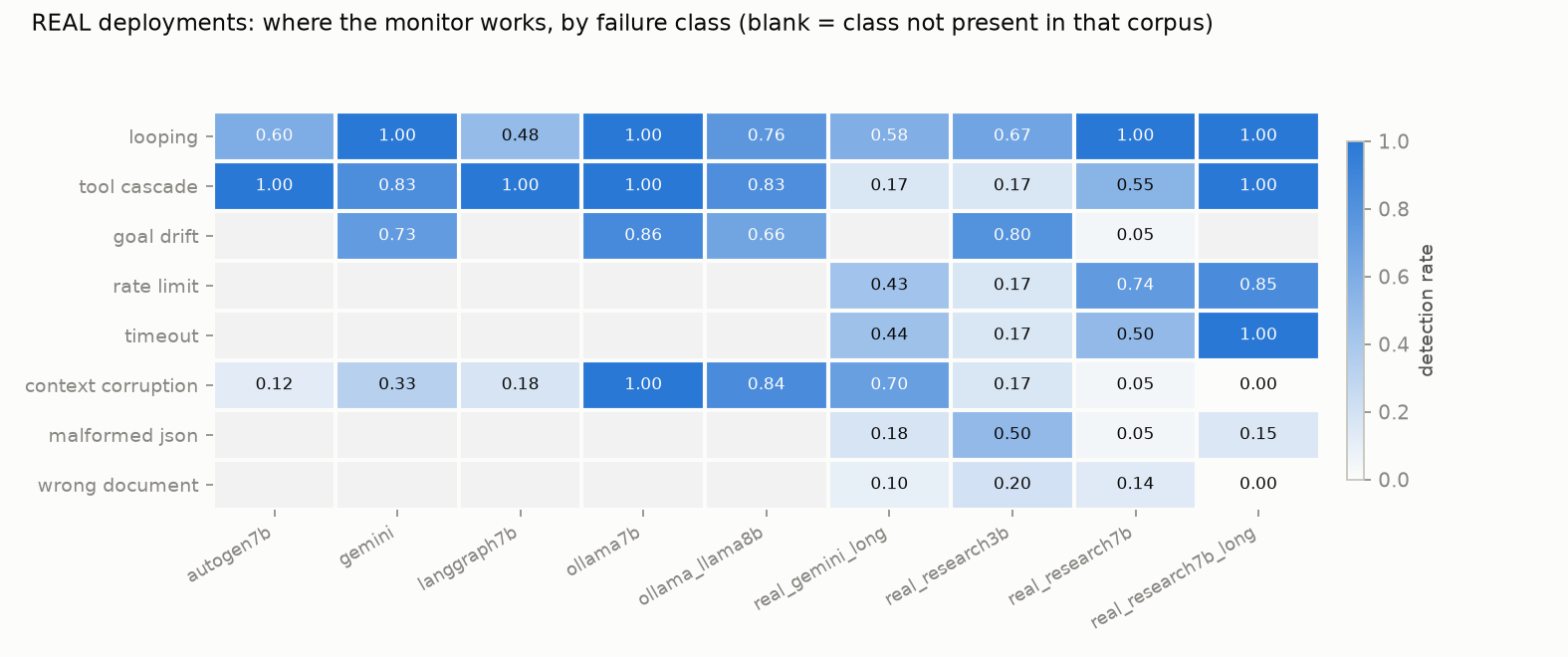}
\caption{\textbf{Real deployments: where the monitor works, by failure class.} Primary-monitor detection rate, nine real corpora, blank where a corpus does not contain that class (absent and undetected are different claims). This is the honest coverage picture the simulator cannot give: looping is caught almost everywhere (0.48--1.00) and tool cascade widely but not uniformly (0.17--1.00), \emph{goal drift} is caught wherever it is present (0.66--0.86 on four of five corpora), and the genuine weak spots are the content-corruption classes --- \texttt{wrong\_document} 0.00--0.20 and \texttt{malformed\_json} 0.05--0.50 --- which is precisely what the grounding channel (\S\ref{sec:grounding}) exists to address.}
\label{fig:coverage}
\end{figure}

Figure~\ref{fig:traces} shows the primary monitor on real streams: healthy runs stay two orders of magnitude below the alarm threshold, and each injected class alarms one step after its verified onset. Figure~\ref{fig:coverage} gives the coverage picture across all nine real corpora, and it is the honest one. Looping is caught almost everywhere (0.48--1.00); tool cascade is caught widely but not uniformly (0.17--1.00, at or above 0.55 on seven of nine corpora and 0.17 on two); goal drift is caught wherever it is present (0.66--0.86 on four of five corpora); and the genuine weak spots are the content-corruption classes, \texttt{wrong\_document} (0.00--0.20) and \texttt{malformed\_json} (0.05--0.50) --- which is what motivates the grounding channel of \S\ref{sec:grounding}, not a post-hoc rationalisation of it.

Two structural findings shape everything that follows. First, monitors do not transfer across deployments without recalibration: calibrated on qwen2.5:7b and deployed on llama3.1:8b over a matched task, tool and injector plan, the primary monitor sits at chance (AUROC 0.527, healthy FA 0.75), while the \emph{same} target recalibrated on itself reaches 0.885 --- so this is a property of the calibration, not of a hard corpus. Second, the memoryless delta-Mahalanobis wins outright on the short-episode datasets (\texttt{real\_research7b} 0.848 vs 0.777, \texttt{real\_research3b} 0.665 vs 0.556), which the horizon diagnosis explains: over 1,002 injected episodes the ESN--Mahalanobis detection gap is \textbf{+0.09} when the post-onset horizon is $\le$3 steps, \textbf{+0.14} at 4--8 and \textbf{+0.40} at $\ge$9 (correlation +0.25). The reservoir needs post-onset steps to integrate evidence: where it has them it wins decisively, and where it does not its margin collapses to near-nothing and the memoryless distance becomes competitive. Note the direction --- averaged over episodes the ESN does not actually \emph{lose} in the short-horizon band; it loses at the dataset level, where the short-episode corpora are also the ones whose healthy nulls are smallest.

\begin{figure}[tbp]
\centering
\includegraphics[width=0.86\linewidth]{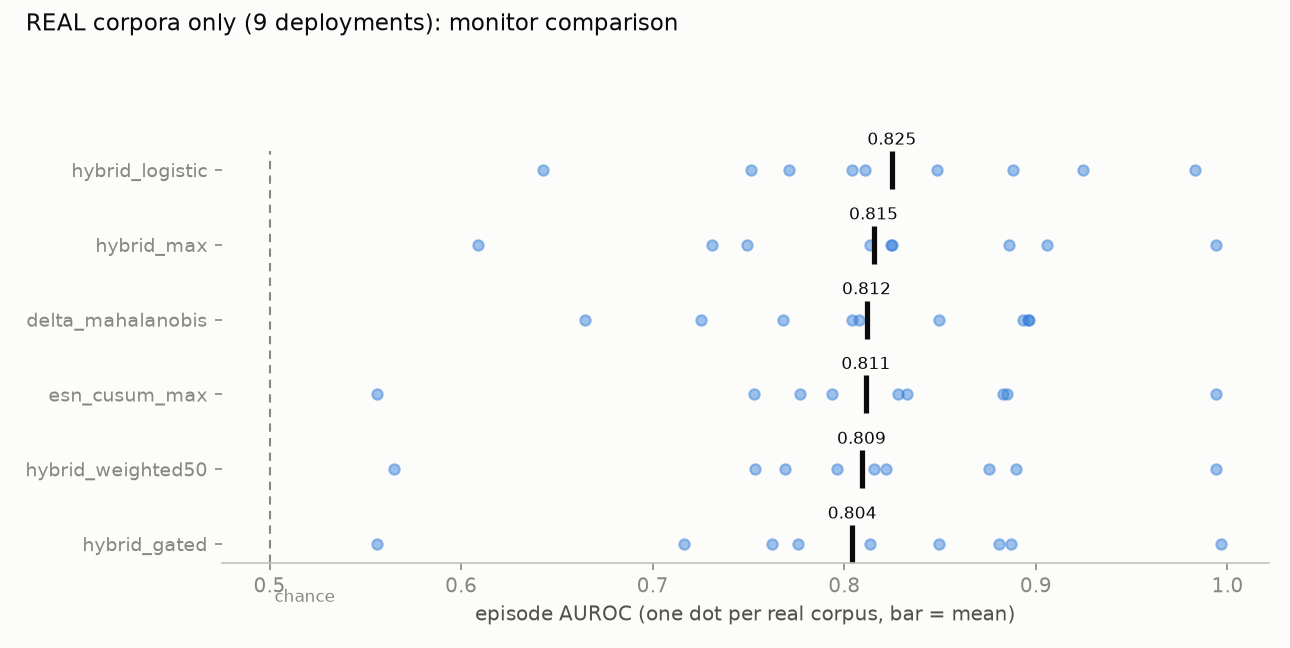}
\caption{\textbf{Monitor comparison on the real corpora only.} Episode AUROC, one dot per real deployment and a bar at the mean; the simulator is excluded. The spread is the point: several per-dataset orderings that look decisive are ties once power is accounted for, so the defensible comparison is the pooled one rather than any single deployment.}
\label{fig:benchmark}
\end{figure}

\textbf{Breadth: a second model family and a commercial API.} Recalibrated on itself, a llama3.1:8b corpus collected on an identical task, tool and injector plan reaches AUROC \textbf{0.885}, so the monitors are not specific to one model family. A lengthened Gemini corpus (125 episodes, 65 positives) reaches \textbf{0.794}: the approach works on a commercial API too. That corpus is gemini-2.5-flash on the same long research task, tool roster and injector plan as the qwen long corpus, so the provider is the only thing that differs. Its lower detection at a matched false-alarm budget (0.38 against 0.57) is a \emph{horizon} effect, not a telemetry one --- its positives are far shorter after onset (median 4 steps against 8), and at matched horizon the two agree (0.53 against 0.56). Ablating the token-surprisal channel inside four corpora that carry it moves the ESN by AUROC $+0.000$, so the deployment requirement is structured tool results, not logprob access. Figure~\ref{fig:benchmark} compares all monitors across these real deployments and is deliberately unflattering: the per-dataset spread is wide enough that several orderings which look decisive are ties once power is accounted for, so only the pooled comparison is defensible.

\begin{figure}[tbp]
\centering
\includegraphics[width=0.9\linewidth]{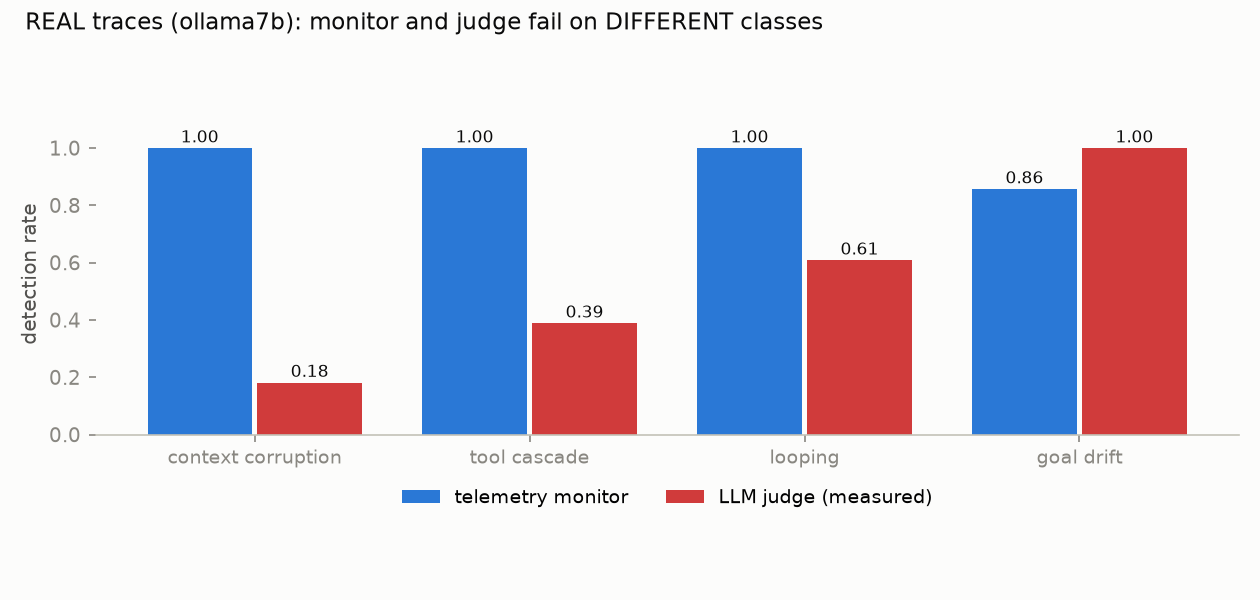}
\caption{\textbf{Monitor and judge fail on different classes.} Both series are measured on the same real corpus: the judge rates from a live gemini-2.5-flash run on a labelled subset (161 distinct prompts), the monitor rates from the primary monitor on that corpus. The judge is perfect on goal drift and nearly blind on context corruption (0.18); the monitor is the reverse on context corruption (1.00). This is the empirical case for escalation as a \emph{complementary layer} rather than a cheaper approximation of the judge.}
\label{fig:complementarity}
\end{figure}

\textbf{The judge is measured, not assumed.} A real gemini-2.5-flash judge on a labelled subset of 161 distinct prompts scores p\_detect \textbf{0.548} (95\% CI 0.44--0.65) and p\_false \textbf{0.052}, against the 0.90/0.02 the escalation analysis assumes. Substituting the measured rates across all five master seeds leaves the call saving intact but drops detection recovery from 82\% to \textbf{43\%}, supported at zero of five seeds. Figure~\ref{fig:complementarity} shows why the layer is still worth having: judge and monitor fail on \emph{different} classes --- the judge is perfect on goal drift and nearly blind on context corruption (0.18), where the monitor is the reverse (1.00).

\subsection{External corpora: two benchmarks we did not build}
\label{sec:external}

On \textbf{AFTraj-2K} \cite{zhang2026agentforesight} --- 1,882 trajectories after the same $T\geq4$ filter, 1,111 healthy and 771 failed, imported with no change to the monitors --- the channel-max ESN reaches episode AUROC \textbf{0.745} (best hybrid 0.760) but detects only \textbf{0.048} at the 5\% budget. Ranking transfers to another project's agents and frameworks; the operating point does not. The horizon diagnosis predicts exactly that: at post-onset horizon $\geq9$ the ESN detects \textbf{0.509} against Mahalanobis' 0.170, and at $\leq3$ neither detects anything --- but only 53 of 771 failures (6.9\%) have that much room and 61\% end within three steps of onset. AgentForesight-7B reports 66.44 Exact-F1 here using an LLM auditor at every step; we do not match it and do not claim to. What this supports is that ranking transfers zero-shot at 163\,$\mu$s per step and no model call.

\textbf{ATBench} \cite{atbench2026} (604 of 1,000 trajectories after the same filter; trajectory-level safe/unsafe labels only, so no $\tau$ and no earliness metric) separates the two findings: the ESN reaches AUROC \textbf{0.779} and detects \textbf{0.311} of unsafe runs --- it alarms here, where on AFTraj it could not --- while $\Delta$-Mahalanobis sits at chance (0.457) and drags every 50/50 fusion down with it (0.438--0.463). Per failure mode the split is sharp: over-privileged actions 0.508 and unvalidated tool outputs 0.473, against 0.038 for inaccurate information. What the agent \emph{does} is caught; what it \emph{says} is not --- the content blind spot \S\ref{sec:grounding} addresses, on labels we did not write.


\clearpage
\section{Controlled Study: Detector Mechanics on Simulated Telemetry}
\label{sec:sim}

\textbf{This section is a mechanism study, and no deployment claim rests on it.} The simulator \emph{constructs the detector's input telemetry directly}: it writes the semantic-embedding state, token-uncertainty aggregates, action metadata, latency, output length, and error flags for each step, with no LLM, tokenizer, embedding measurement, or telemetry adapter in the loop. The class-channel signatures are therefore designed in, and what follows measures whether the monitors \emph{recover} the injected structure --- not whether that structure arises from a real model. It earns its place for two reasons the real corpora cannot supply: it provides ground-truth onsets at arbitrary horizons, and it contains classes the real corpora do not (grounding loss). Read it as a cartoon of the mechanism; the evidence is \S\ref{sec:real}.

Its numbers should be read with that caveat attached, and one of them does not survive contact with measurement: the escalation result below assumes a judge we have since measured, and \S\ref{sec:real} reports what happens when the assumption is replaced. Over five master seeds the ESN detects \textbf{0.707 $\pm$ 0.068} of failures at a 4.4\% realized FA with episode AUC \textbf{0.872 $\pm$ 0.015} and a mean lead of 4.6 steps; H1 holds at four of the five seeds and is honestly not supported at one. Label-free confidence is calibrated by the healthy-score null: its healthy stream is uniform to KS $\approx$ 0.12 (fused), and the oracle isotonic posterior (\emph{with} labels) reaches ECE $\approx$ 0.03 \cite{guo2017calibration}. A cost-optimal escalation policy (operating point selected on calibration) recovers \textbf{83\% of judge-every-step detection at 8\% of its judge calls} (master seed; H3b holds at four of five seeds).

\textbf{H3b is conditional on the judge, and the assumption fails.} That judge is a stipulated noisy oracle (p\_detect 0.90, p\_false 0.02). A real gemini-2.5-flash judge measured on a labelled subset of collected episodes (161 distinct prompts) scores p\_detect \textbf{0.548} (95\% CI 0.44--0.65) and p\_false \textbf{0.052} (0.020--0.13) --- both stipulated values lie outside their measured intervals. Substituting the measured rates and re-running all five master seeds with everything else fixed, the \emph{call saving survives} (6\% vs 8\% of judge-every-step calls; cost ratio 0.63 vs 0.61) but detection recovery falls to \textbf{43\%} (range 39--47\%) and H3b is supported at \textbf{0 of 5 seeds}. The escalation architecture earns its cost saving; its detection claim should be read as 43\%, not the 82\% the stipulated judge produced over the same five seeds (83\% at the master seed quoted above). Per class the measured judge is perfect on goal drift (21/21) and nearly blind on context corruption (0.18) --- the class the grounding channel targets --- so judge and monitor fail on different classes.

\section{The Calibrated Hybrid}

\begin{figure}[tbp]
\centering
\includegraphics[width=0.85\linewidth]{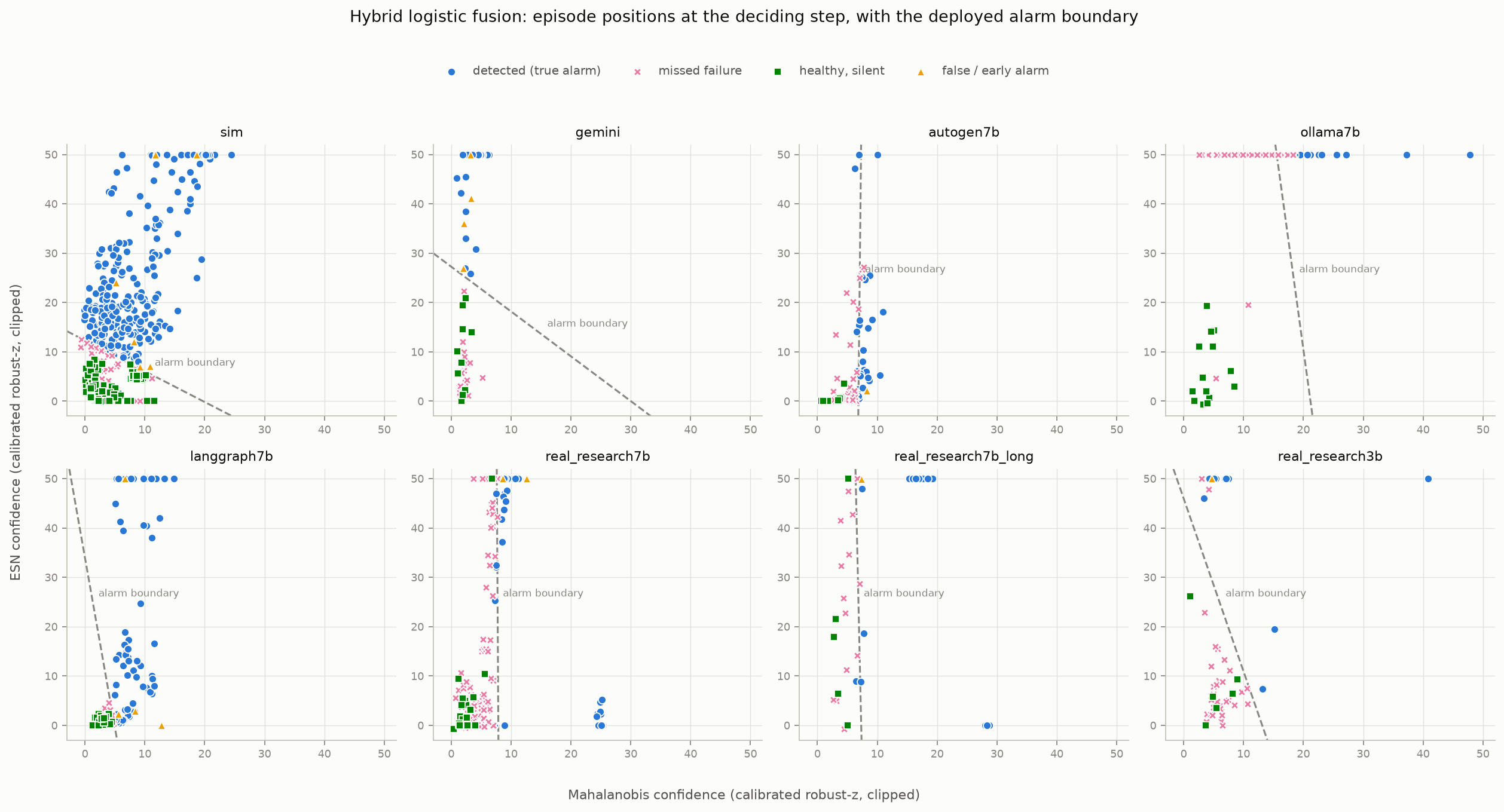}
\caption{Hybrid logistic fusion: episode positions at the deciding step in the (Mahalanobis, ESN) calibrated-confidence plane, one panel per deployment, with the learned alarm boundary. The boundary's orientation is the learned regime: near-vertical (Mahalanobis-driven) on short-horizon real-agent datasets, ESN-driven on the long-horizon simulator.}
\label{fig:hybrid}
\end{figure}

Of 1{,}002 injected episodes, about a third of failures are visible to exactly one detector parent (ESN vs Mahalanobis: 273 ESN-only, 57 Mahalanobis-only, 372 both, 300 neither --- 330 of 1{,}002, or 32.9\%, seen by exactly one). The cross-fit logistic fusion recovers \textbf{100\% / 66\% / 79\%} of the both/ESN-only/Maha-only cells and fires on just 1\% of the cases neither parent flags. Its learned Mahalanobis weight share tracks the regime (from 0.38 on long horizons to 0.99 on short; Figure~\ref{fig:hybrid}). The fusion's advantage is pooled, not local: its grand-mean AUROC (0.826) beats either parent taken alone (ESN 0.802, $\Delta$-Mahalanobis 0.807) because each parent collapses on some dataset and the fusion never does, but measured against whichever parent is better \emph{on that dataset} it is at or below it on 7 of 8 datasets in AUROC (mean $-0.014$) and on all 8 in detection rate (mean $-0.140$). The hybrid is therefore the right default when the deployment regime is unknown, not a dominance result.

\section{Closing the Content Blind Spot}
\label{sec:grounding}

Behavioral and statistical monitors share one blind spot: corruption that changes \textit{data} without changing \textit{behavior}. The $g$ channel adds nine causal content features, including a binary lexical relevance flag that completely separates off-topic retrievals.

\begin{table}[tbp]
\centering
\caption{Detection rate (pooled injected episodes, n=874 over ten datasets,
\texttt{grounding\_diagnosis.csv}). Ungrounded $\rightarrow$ grounded; McNemar
discordant (gained/lost) vs the ungrounded parent.}
\begin{tabular}{lcc}
\toprule
Fusion & Content (n=313) & Behavioral (n=561) \\
\midrule
Content gate & 0.27 $\rightarrow$ \textbf{0.58} (+96/-0) & 0.74 $\rightarrow$ \textbf{0.79} (+27/-0) \\
Joint-budget & 0.27 $\rightarrow$ \textbf{0.45} (+58/-1) & 0.74 $\rightarrow$ 0.73 (+0/-3) \\
Logistic+g (labels) & 0.21 $\rightarrow$ 0.32 (+42/-9) & 0.71 $\rightarrow$ 0.69 (+23/-32) \\
\bottomrule
\end{tabular}
\end{table}

\section{Organic Failures and Conclusion}
\label{sec:organic}

Thirty non-injected high-temperature (0.9) episodes yield 11 organic failures: 7 silent aborts, 3 fabricated counts, 1 ungrounded retrieval blend. Scored by monitors carrying the calibration they were deployed with, and not refit on this data, the behavioural monitors catch only \textbf{1 of the 3} fabrications and rank the whole organic failure set at/below chance (within-organic AUROC \textbf{0.31--0.42}); a temperature-matched null still false-alarms on \textbf{36\%} of healthy episodes, because the organic runs were sampled at temperature 0.9. Detections at that false-alarm rate are not deployable evidence, and we report them as such. \textbf{Omission is invisible to telemetry but trivially fixable} (the five-line completion check catches 7/7 silent aborts with 0 false positives across these 30 episodes), and \textbf{plausible-value corruption remains undetectable} without an external reference.

A \textbf{pre-registered replication} (design and analysis committed before data; $n=55$ episodes at temperature 0.9) built the null cross-fit from the healthy subset of the same-temperature runs, but is \textbf{explicitly underpowered}: only 2 candidate hallucinations arise (pre-registered minimum 10), at a 36\% healthy false-alarm rate, so it makes \textbf{no detection claim}. These well-aligned agent models rarely invent figures. Fabrication detection is therefore delivered not by a statistical monitor but by the mechanism that matches the failure: a deterministic per-step \textbf{numeric-grounding verifier} --- every monetary figure the agent asserts must trace to a tool result it actually received, or an arithmetic combination thereof --- which needs no ground-truth answer, runs online, flagged 0 false positives on the 25 healthy episodes, and catches an inserted ungrounded figure at the step it appears. The statistical monitors were never the right tool for this class; measuring that, and shipping the right tool, is the result.

Observable step telemetry covers most of the agent-failure space using microsecond-scale monitors --- with failure classes it cannot cover delegated to mechanisms that can (completion check, grounding verifier, escalation to a judge) --- providing a vital first line of defense before escalating to full LLM judges.

\section{Deterministic Verification and Repair}
\label{sec:verify}

The monitors above need a healthy null per deployment and still spend part of a
false-alarm budget. A check needs neither. \texttt{total\_consistency}
recomputes a run's stated total from the tool results \emph{that run actually
received}; \texttt{required\_coverage} confirms every call the task requires was
made; \texttt{tool\_contract} asks whether a result matched any shape its tool can
return, and so reports at the step the result arrives rather than at the answer.
None reads the hidden world the task was generated from, so all three are
deployable as-is.

\begin{table}[h]
\centering
\caption{Checks versus the behavioural monitor, same episodes and same objective
labels (\texttt{verification\_vs\_monitor.csv}). Recall is comparable at the
served temperature; the difference is precision.}
\label{tab:verify}
\begin{tabular}{lcc}
\toprule
 & checks & monitor \\
\midrule
failures caught, $T{=}0.2$ (served) & 60\% (\textbf{96\%} with coverage) & 54\% \\
false positives, $T{=}0.2$ & \textbf{0/63 = 0\%} & 11/63 = 17\% \\
failures caught, $T{=}0.9$ & 65\% (\textbf{96\%} with coverage) & 40\% \\
false positives, $T{=}0.9$ & \textbf{0/38 = 0\%} & 6/38 = 16\% \\
\bottomrule
\end{tabular}
\end{table}

The checks were written by inspecting failures in the serving arm, so that arm
cannot also be their test set. A further 120 episodes at disjoint task seeds
(40000+, zero overlap) were collected afterwards and scored with the checks
frozen: 54\% caught by totals, \textbf{93\%} with coverage, arithmetic errors
36/36, and \textbf{0 of 64} false positives --- the overall figure drops by the
margin a genuine held-out test should cost, driven by the small
\texttt{hallucinated} class (4/8). A llama3.1:8b arm on the \emph{same} 120 task
seeds, with nothing retuned, catches \textbf{110 of 110} failures at \textbf{0 of
10} false positives. On the provoked corpus the checks catch \textbf{26 of 26}
fabrications; that corpus cannot score a one-class monitor at all, because
provoking enough fabrication leaves 2 healthy episodes against the 15 a null
needs. Scored across every labelled corpus, \texttt{tool\_contract} trips on
\textbf{0 of 1825} healthy episodes, flags 46\% of \texttt{context\_corruption}
and 44\% of \texttt{looping} and 0\% of every other injected class, and where it
fires is immediate: \textbf{215 of 218} flagged episodes within one step of onset.

\textbf{Repair.} Every flagged episode is rolled back to the same checkpoint and
re-run under each repair rung, paired on the identical prefix and task; the
rollback is real (a committed trace plus its seed rebuilds the conversation at
step $k$, and every step after is a fresh model call), and success is graded by
the study oracle, which the repair prompt never sees. Each cell is the mean of
three independent repeats over $n=55$ genuinely-wrong episodes.

\begin{table}[h]
\centering
\caption{Repair rungs against a resampling control (\texttt{repair\_policies.csv}).
Asking for a re-check is what works, and naming the failing check works best.}
\label{tab:repair}
\begin{tabular}{lccc}
\toprule
rung & recovery & vs \texttt{resample} & calls/recovery \\
\midrule
\texttt{none} --- untouched & 0\% & --- & --- \\
\texttt{resample} --- rollback + fresh sample & 16\% & \emph{control} & 14.7 \\
\textbf{\texttt{located}} --- + which check failed, no values & \textbf{45\%} & \textbf{$p=0.0005$} & 6.4 \\
\texttt{generic} --- + a re-check instruction & 36\% & $p=0.0347$ & \textbf{5.8} \\
\texttt{specific} --- + the finding, with values & 36\% & $p=0.0192$ & 8.1 \\
\texttt{recompute} --- + use the calculator & 28\% & $p=0.17$ (n.s.) & 7.2 \\
\texttt{adaptive} & 21\% & $p=0.61$ (n.s.) & 10.6 \\
\bottomrule
\end{tabular}
\end{table}

Net over all 120 episodes, charging each policy for any correct run it broke,
\texttt{located} lifts task success from \textbf{52\% to 73\%}: 25 failures
recovered and \textbf{zero} correct runs broken, because the checks flagged no
already-correct episode. Retry luck is controlled for --- plain resampling alone
recovers 16\%, so only the margin above that is credited to the repair.
\textbf{Five rungs are tested against that one control, and only the headline
survives correcting for it}: at Bonferroni $\alpha=0.05/5=0.01$,
\texttt{located} ($p=0.0005$) holds, while \texttt{generic} ($p=0.0347$) and
\texttt{specific} ($p=0.0192$) do not, and should be read as suggestive rather
than established. This sharpens rather than weakens the finding --- naming the
failing check is the rung that survives. Two
negative results sit inside this table and are kept rather than dropped.
\texttt{recompute} routes the step to a calculator the agent is already holding,
which should fix the dominant arithmetic failure, and does not beat retry luck at
this sample size ($p=0.17$). And supplying the recomputed answer buys nothing:
26 of 55 \texttt{specific} hints contain the correct total outright, while
\texttt{located} states no value at all and recovers at least as much --- so the
recovery is not coming from being handed the answer.

Repair coverage is real but partial, and the shape of the gap is measured. Over
five injection classes $\times$ five task seeds with halting off
(\texttt{alarm\_repair.csv}, $n=25$ live episodes) every behavioural alarm is
followed by a repair attempt --- 21 of 21, and 18 of 18 in an independent run of
the same matrix --- but \texttt{goal\_drift} is the only class a retry fixes (4
of 5). Per-class alarm rates move between runs of this live matrix, so they are
a sample rather than a constant. Where the tool layer itself is broken a retry fetches the same broken result:
\texttt{tool\_cascade} and \texttt{looping} escalate rather than recover, and the
value of the intervention there is ending the episode fast --- a loop trap exits
at exactly 10 steps in 5 of 5 runs, against 30 before the circuit breaker.
\texttt{grounding\_loss} never alarms behaviourally (0 of 5) and is the grounding
check's class by design. A contract violation is escalated, never repaired.

\section{Limitations}
\label{sec:limits}

We state the boundaries as measured, not as caveats.

\textbf{(1) No cross-deployment transfer.} The healthy null must be collected
under the exact serving distribution --- temperature, toolset, probe pattern,
latency regime. Across families on a matched plan (qwen2.5:7b
$\rightarrow$ llama3.1:8b) transferred monitors sit at chance (AUROC 0.527)
where the same target recalibrated on itself reaches 0.885, so the failure is
the calibration, not a hard target. Within a family (7b $\rightarrow$ 3b) it
also lands at chance, but reads less cleanly: the 3b corpus only reaches 0.556
recalibrated, so ``hard target'' survives as an explanation there.

\textbf{(2) Slow goal drift} evades every per-step-surprise monitor tested;
only trajectory self-consistency sees it, weakly. The limit is the \emph{rate}
of change, not the class --- abrupt goal changes are caught at 0.66--0.86 on
four of five real corpora (Figure~\ref{fig:coverage}).

\textbf{(3) Content coverage is bounded by telemetry completeness.} Results not
passed through the telemetry channel are invisible, and terse results carry too
little mass to detect garbling. The bound is on \emph{structured tool results},
not logprobs: ablating the token-surprisal channel moves the ESN by AUROC
$+0.000$ and detection $+0.002$, so a provider that withholds them loses almost
nothing.

\textbf{(4) Plausible-value corruption} --- a wrong-but-well-formed number ---
is undetectable from telemetry by construction and needs an external reference.

\textbf{(5) Fabrication claims are limited by base rate.} The objective
labeller flags 9 of 175 organic episodes as hallucinated and only \textbf{2 as
fabricated inputs}, the class the grounding verifier targets --- both below the
pre-registered minimum of 10, so no detection claim is made in either
direction. Under provocation the class becomes testable and the grounding verifier
catches \textbf{0.55} of 11 ungrounded-input fabrications at 0 false positives
on the 9 healthy episodes of that corpus: specific, about half sensitive, and
only under provocation.
That 0.55 and the \textbf{26 of 26} in \S\ref{sec:verify} are different layers
on different denominators and not in tension: the deterministic checks catch
all 26 provoked hallucinations by recomputing the total, while 0.55 is the
grounding verifier alone on the narrower ungrounded-\emph{input} subset.

\textbf{(6) Scope.} Mock-tool and research-loop tasks, two local model families
and one commercial API; wall-clock latency features are machine-specific and
are excluded from the shipped configuration.

\section*{Appendix}
\sloppy

\subsection*{Artifact availability}

All code, collected traces, result tables and figures are released together. Nothing in this paper is computed from data held back.

\textbf{Licence scope.} This manuscript's licence, and the MIT licence on the repository, cover only what this project wrote: the source, the trace format, the result tables and figures, and this text. They do not relicense material this project merely recorded or called, which keeps its own terms --- \texttt{qwen2.5} output under Apache-2.0, \texttt{llama3.1:8b} output under the Llama 3.1 Community License (\textbf{Built with Llama}), \texttt{gemini-2.5-flash} output under the Gemini API Additional Terms, and replayed weather and search results under Open-Meteo CC~BY~4.0 and Wikipedia CC~BY-SA~4.0. The per-corpus breakdown is in the repository's \texttt{DATA\_CARD.md}; the external benchmarks of \S\ref{sec:external} are downloaded rather than redistributed, and remain their authors'.

\begin{itemize}
\item Code, traces and results: \url{https://github.com/sunnydubey1111/agent-trajectory-sentinel}
\item The corpus as a loadable dataset: \url{https://huggingface.co/datasets/sunnydubey1111/agent-trajectory-sentinel}
\item An interactive replay of the monitor scoring real runs: \url{https://huggingface.co/spaces/sunnydubey1111/agent-trajectory-sentinel-demo}
\item A recorded walkthrough of the method and the live demo: \url{https://youtu.be/a05n_000klE}
\end{itemize}

\noindent The layout a reader needs is:

\begin{itemize}
\item \texttt{results/tables/} --- every CSV and JSON named in Table~\ref{tab:provenance}. A claim in the text and the file named beside it are the same numbers; the file is the source.
\item \texttt{results/figures/} --- the figures in this paper, regenerated by \texttt{py -m derail.experiments.plots}.
\item \texttt{traces/} --- the agent episodes themselves, one JSONL file per episode with a \texttt{manifest.json} per corpus recording each episode's model, label, verified onset and checksum.
\item \texttt{BASELINE\_MANIFEST.json} --- a SHA-256 for every source file, result artifact and trace in the repository. It is what makes the provenance in Table~\ref{tab:provenance} checkable rather than asserted: \texttt{py -m devtools.artifact\_manifest -{}-check} recomputes every hash and reports any file that differs from the state these results were produced in.
\end{itemize}

Two further checks ship with the code. \texttt{py -m devtools.behavior\_snapshot -{}-check} re-runs the study at a fixed seed and compares every value against a stored baseline, so a change in behaviour cannot pass unnoticed; and \texttt{py -m pytest} runs the full suite, including the slow tests that exercise real tools. A container definition and a pinned lock file are included for a network-free CPU reproduction of the synthetic study.

\begin{table}[H]
\centering
\caption{Provenance map: each claim to the artifact it is computed from. Every
file listed is committed, and \texttt{BASELINE\_MANIFEST.json} records a SHA-256
for it, so a reader can verify that the number in the text came from the file in
the repository.}
\label{tab:provenance}
\scriptsize
\begin{tabular}{p{2.9cm} p{5.0cm} p{4.3cm}}
\toprule
Claim & Artifact & Regenerate with \\
\midrule
Real per-class coverage & \texttt{l7b\_per\_class.csv} & \texttt{run\_hybrid\_study} \\
Cross-family transfer & \texttt{model\_transfer\_family.csv} & \texttt{run\_model\_transfer} \\
gemini-2.5-flash corpus & \texttt{gemini\_long\_*.csv} & \texttt{run\_hybrid\_study} \\
Measured judge & \texttt{judge\_calibration\_summary.json} & \texttt{run\_judge\_calibration} \\
Judge consequence for H3b & \texttt{judge\_sensitivity.csv} & \texttt{judge\_sensitivity} \\
Telemetry-channel ablation & \texttt{telemetry\_dependence.csv} & \texttt{telemetry\_dependence} \\
Recalibration cost & \texttt{recalibration\_cost.csv} & \texttt{recalibration\_cost} \\
Statistical power & \texttt{power\_analysis*.csv} & \texttt{power\_analysis} \\
Adversarial limit & \texttt{adversarial\_evasion.csv}, \texttt{tamper\_check.csv} & \texttt{tamper\_check} \\
Organic + provoked fabrication & \texttt{organic\_hallucination*.csv}, \texttt{provoked\_fabrication.csv} & \texttt{score\_provoked\_fabrication} \\
Simulator study & \texttt{multiseed*.csv}, \texttt{h1\_*}, \texttt{h3\_*} & \texttt{run\_multiseed} \\
\bottomrule
\end{tabular}
\end{table}

\bibliographystyle{unsrt}
\bibliography{references}

\begin{thebibliography}{10}

\bibitem{zheng2023judging}
Lianmin Zheng, Wei-Lin Chiang, Ying Sheng, Siyuan Zhuang, Zhanghao Wu, Yonghao
  Zhuang, Zi~Lin, Zhuohan Li, Dacheng Li, Eric~P. Xing, Hao Zhang, Joseph~E.
  Gonzalez, and Ion Stoica.
\newblock Judging {LLM}-as-a-judge with {MT-Bench} and {Chatbot Arena}.
\newblock In {\em Advances in Neural Information Processing Systems (NeurIPS)
  Datasets and Benchmarks Track}, 2023.

\bibitem{page1954cusum}
E.~S. Page.
\newblock Continuous inspection schemes.
\newblock {\em Biometrika}, 41(1/2):100--115, 1954.

\bibitem{basseville1993detection}
Mich{\`e}le Basseville and Igor~V. Nikiforov.
\newblock {\em Detection of Abrupt Changes: Theory and Application}.
\newblock Prentice Hall, 1993.

\bibitem{chandola2009anomaly}
Varun Chandola, Arindam Banerjee, and Vipin Kumar.
\newblock Anomaly detection: A survey.
\newblock {\em ACM Computing Surveys}, 41(3):1--58, 2009.

\bibitem{lee2018simple}
Kimin Lee, Kibok Lee, Honglak Lee, and Jinwoo Shin.
\newblock A simple unified framework for detecting out-of-distribution samples
  and adversarial attacks.
\newblock In {\em Advances in Neural Information Processing Systems (NeurIPS)},
  2018.

\bibitem{hendrycks2017baseline}
Dan Hendrycks and Kevin Gimpel.
\newblock A baseline for detecting misclassified and out-of-distribution
  examples in neural networks.
\newblock In {\em International Conference on Learning Representations (ICLR)},
  2017.

\bibitem{liu2008isolation}
Fei~Tony Liu, Kai~Ming Ting, and Zhi-Hua Zhou.
\newblock Isolation forest.
\newblock In {\em IEEE International Conference on Data Mining (ICDM)}, pages
  413--422, 2008.

\bibitem{jaeger2001echo}
Herbert Jaeger.
\newblock The ``echo state'' approach to analysing and training recurrent
  neural networks.
\newblock Technical Report 148, GMD -- German National Research Institute for
  Computer Science, 2001.

\bibitem{lukosevicius2009reservoir}
Mantas Luko{\v{s}}evi{\v{c}}ius and Herbert Jaeger.
\newblock Reservoir computing approaches to recurrent neural network training.
\newblock {\em Computer Science Review}, 3(3):127--149, 2009.

\bibitem{guo2017calibration}
Chuan Guo, Geoff Pleiss, Yu~Sun, and Kilian~Q. Weinberger.
\newblock On calibration of modern neural networks.
\newblock In {\em International Conference on Machine Learning (ICML)}, 2017.

\bibitem{yao2023react}
Shunyu Yao, Jeffrey Zhao, Dian Yu, Nan Du, Izhak Shafran, Karthik Narasimhan,
  and Yuan Cao.
\newblock {ReAct}: Synergizing reasoning and acting in language models.
\newblock In {\em International Conference on Learning Representations (ICLR)},
  2023.

\bibitem{wu2023autogen}
Qingyun Wu, Gagan Bansal, Jieyu Zhang, Yiran Wu, Beibin Li, Erkang Zhu,
  Li~Jiang, Xiaoyun Zhang, Shaokun Zhang, Jiale Liu, Ahmed~Hassan Awadallah,
  Ryen~W. White, Doug Burger, and Chi Wang.
\newblock {AutoGen}: Enabling next-gen {LLM} applications via multi-agent
  conversation.
\newblock {\em arXiv preprint arXiv:2308.08155}, 2023.

\bibitem{liu2024agentbench}
Xiao Liu, Hao Yu, Hanchen Zhang, Yifan Xu, Xuanyu Lei, Hanyu Lai, Yu~Gu,
  Hangliang Ding, Kaiwen Men, Kejuan Yang, Shudan Zhang, Xiang Deng, Aohan
  Zeng, Zhengxiao Du, Chenhui Zhang, Sheng Shen, Tianjun Zhang, Yu~Su, Huan
  Sun, Minlie Huang, Yuxiao Dong, and Jie Tang.
\newblock {AgentBench}: Evaluating {LLMs} as agents.
\newblock In {\em International Conference on Learning Representations (ICLR)},
  2024.

\bibitem{cemri2025why}
Mert Cemri, Melissa~Z. Pan, Shuyi Yang, Lakshya~A. Agrawal, Bhavya Chopra,
  Rishabh Tiwari, Kurt Keutzer, Aditya Parameswaran, Dan Klein, Kannan
  Ramchandran, Matei Zaharia, Joseph~E. Gonzalez, and Ion Stoica.
\newblock Why do multi-agent {LLM} systems fail?
\newblock {\em arXiv preprint arXiv:2503.13657}, 2025.

\bibitem{zhang2025which}
Shaokun Zhang, Ming Yin, Jieyu Zhang, Jiale Liu, Zhiguang Han, Jingyang Zhang,
  Beibin Li, Chi Wang, Huazheng Wang, Yiran Chen, and Qingyun Wu.
\newblock Which agent causes task failures and when? on automated failure
  attribution of {LLM} multi-agent systems.
\newblock {\em arXiv preprint arXiv:2505.00212}, 2025.

\bibitem{zhang2026agentforesight}
Boxuan Zhang, Jianing Zhu, Zeru Shi, Dongfang Liu, and Ruixiang Tang.
\newblock {AgentForesight}: Online auditing for early failure prediction in
  multi-agent systems.
\newblock {\em arXiv preprint arXiv:2605.08715}, 2026.

\bibitem{baidya2026sparse}
Avinash Baidya, Xinran Liang, Ruocheng Guo, Xiang Gao, and Kamalika Das.
\newblock When evidence is sparse: Weakly supervised early failure alerting in
  dialogs and {LLM}-agent trajectories.
\newblock {\em arXiv preprint arXiv:2606.05414}, 2026.

\bibitem{wang2025agentspec}
Haoyu Wang, Christopher~M. Poskitt, and Jun Sun.
\newblock {AgentSpec}: Customizable runtime enforcement for safe and reliable
  {LLM} agents.
\newblock In {\em IEEE/ACM International Conference on Software Engineering
  (ICSE)}, 2026.
\newblock arXiv:2503.18666, 2025.

\bibitem{wang2025probguard}
Haoyu Wang, Christopher~M. Poskitt, Jiali Wei, and Jun Sun.
\newblock {ProbGuard}: Probabilistic runtime monitoring for {LLM} agent safety.
\newblock {\em arXiv preprint arXiv:2508.00500}, 2025.

\bibitem{huang2026prefixguard}
Xinmiao Huang, Jinwei Hu, Rajarshi Roy, Changshun Wu, Yi~Dong, and Xiaowei
  Huang.
\newblock {PrefixGuard}: From {LLM}-agent traces to online failure-warning
  monitors.
\newblock {\em arXiv preprint arXiv:2605.06455}, 2026.

\bibitem{advani2026trajectoryguard}
Laksh Advani.
\newblock Trajectory guard: A lightweight, sequence-aware model for real-time
  anomaly detection in agentic {AI}.
\newblock {\em arXiv preprint arXiv:2601.00516}, 2026.

\bibitem{zhao2026agenttether}
Chenyu Zhao, Shenglin Zhang, Wenwei Gu, Yongqian Sun, Dan Pei, Chetan Bansal,
  Saravan Rajmohan, and Minghua Ma.
\newblock {AgentTether}: Graph-guided diagnosis and runtime intervention for
  reliable {LLM} agent operation.
\newblock {\em arXiv preprint arXiv:2607.06273}, 2026.

\bibitem{gu2024survey}
Jiawei Gu, Xuhui Jiang, Zhichao Shi, Hexiang Tan, Xuehao Zhai, Chengjin Xu, Wei
  Li, Yinghan Shen, Shengjie Ma, Honghao Liu, Saizhuo Wang, Kun Zhang, Yuanzhuo
  Wang, Wen Gao, Lionel Ni, and Jian Guo.
\newblock A survey on {LLM}-as-a-judge.
\newblock {\em arXiv preprint arXiv:2411.15594}, 2024.

\bibitem{shi2024judging}
Lin Shi, Chiyu Ma, Wenhua Liang, Xingjian Diao, Weicheng Ma, and Soroush
  Vosoughi.
\newblock Judging the judges: A systematic study of position bias in
  {LLM}-as-a-judge.
\newblock In {\em Conference of the Asia-Pacific Chapter of the Association for
  Computational Linguistics (AACL-IJCNLP)}, 2025.
\newblock arXiv:2406.07791, 2024.

\bibitem{ji2023survey}
Ziwei Ji, Nayeon Lee, Rita Frieske, Tiezheng Yu, Dan Su, Yan Xu, Etsuko Ishii,
  Ye~Jin Bang, Andrea Madotto, and Pascale Fung.
\newblock Survey of hallucination in natural language generation.
\newblock {\em ACM Computing Surveys}, 55(12):1--38, 2023.

\bibitem{lin2025hallucination}
Xixun Lin, Yucheng Ning, Jingwen Zhang, Yan Dong, Yilong Liu, Yongxuan Wu,
  Xiaohua Qi, Nan Sun, Yanmin Shang, Kun Wang, Pengfei Cao, Qingyue Wang, Lixin
  Zou, Xu~Chen, Chuan Zhou, Jia Wu, Peng Zhang, Qingsong Wen, Shirui Pan, Bin
  Wang, Yanan Cao, Kai Chen, Songlin Hu, and Li~Guo.
\newblock {LLM}-based agents suffer from hallucinations: A survey of taxonomy,
  methods, and directions.
\newblock {\em arXiv preprint arXiv:2509.18970}, 2025.

\bibitem{manakul2023selfcheckgpt}
Potsawee Manakul, Adian Liusie, and Mark J.~F. Gales.
\newblock {SelfCheckGPT}: Zero-resource black-box hallucination detection for
  generative large language models.
\newblock In {\em Conference on Empirical Methods in Natural Language
  Processing (EMNLP)}, 2023.

\bibitem{farquhar2024detecting}
Sebastian Farquhar, Jannik Kossen, Lorenz Kuhn, and Yarin Gal.
\newblock Detecting hallucinations in large language models using semantic
  entropy.
\newblock {\em Nature}, 630:625--630, 2024.

\bibitem{kossen2024semantic}
Jannik Kossen, Jiatong Han, Muhammed Razzak, Lisa Schut, Shreshth Malik, and
  Yarin Gal.
\newblock Semantic entropy probes: Robust and cheap hallucination detection in
  {LLMs}.
\newblock {\em arXiv preprint arXiv:2406.15927}, 2024.

\bibitem{goldowskydill2025detecting}
Nicholas Goldowsky-Dill, Bilal Chughtai, Stefan Heimersheim, and Marius
  Hobbhahn.
\newblock Detecting strategic deception with linear probes.
\newblock In {\em International Conference on Machine Learning (ICML)}, volume
  267 of {\em PMLR}, pages 19755--19786, 2025.

\bibitem{ruan2026doomed}
Kai Ruan, Zihe Huang, Ziqi Zhou, Qianshan Wei, Xuan Wang, and Hao Sun.
\newblock Doomed from the start: Early abort of {LLM} agent episodes via a
  recall-controlled probe cascade.
\newblock {\em arXiv preprint arXiv:2607.06503}, 2026.

\bibitem{atbench2026}
{Shanghai AI Laboratory}.
\newblock {ATBench}: A diverse and realistic agent trajectory benchmark for
  long-horizon agent safety.
\newblock {\em arXiv preprint arXiv:2604.02022}, 2026.

\end{thebibliography}

\end{document}